\documentclass[journal,comsoc]{IEEEtran}
\usepackage[T1]{fontenc}
\usepackage[pdftex]{graphicx}
\usepackage{bm}
\usepackage{mathrsfs}
\usepackage{amsthm}
\usepackage{amsmath}
\usepackage{amssymb}
\usepackage{algorithm}
\usepackage{algorithmic}
\usepackage{array}
\usepackage{bm}
\usepackage{booktabs}
\usepackage{caption3}
\usepackage{url}
\usepackage{multirow}
\usepackage{xcolor}
\usepackage{float}
\usepackage[caption=false,font=footnotesize]{subfig}
\newcommand{\myrevisedcolor}[1]{\textcolor{black}{{}#1}}

\ifCLASSINFOpdf
\else
\fi
\usepackage{amsmath}
\usepackage[cmintegrals]{newtxmath}

\begin{document}
%
\title{Discriminative Supervised Subspace Learning for Cross-modal Retrieval}
%
%
%

\author{Haoming~Zhang,
        Xiao-Jun~Wu*,
        Tianyang Xu and Donglin Zhang
        \thanks{This work was supported in part by NSFC (61672265, U1836218, 62020106012)}
\thanks{H. Zhang, X.-J. Wu, T. Xu and D. Zhang are with the School of Artificial Intelligence and Computer Science, Jiangnan University, Wuxi, P.R. China. (e-mail: wu\_xiaojun@jiangnan.edu.cn) (*Corresponding author is Xiao-Jun Wu)}
}

%


\maketitle

\begin{abstract}
Nowadays the measure between heterogeneous data is still an open problem for cross-modal retrieval. The core of cross-modal retrieval is how to measure the similarity between different types of data. Many approaches have been developed to solve the problem. As one of the mainstream, approaches based on subspace learning pay attention to learning a common subspace where the similarity among multi-modal data can be measured directly. However, many of the existing approaches only focus on learning a latent subspace. They ignore the full use of discriminative information so that the semantically structural information is not well preserved. Therefore satisfactory results can not be achieved as expected. We in this paper propose a discriminative supervised subspace learning for cross-modal retrieval(DS²L), to make full use of discriminative information and better preserve the semantically structural information. Specifically, we first construct a shared semantic graph to preserve the semantic structure within each modality. Subsequently, the Hilbert-Schmidt Independence Criterion(HSIC) is introduced to preserve the consistence between feature-similarity and semantic-similarity of samples. Thirdly, we introduce a similarity preservation term, thus our model can compensate for the shortcomings of insufficient use of discriminative data and better preserve the semantically structural information within each modality.    \myrevisedcolor{The experimental results obtained on three well-known benchmark datasets demonstrate the effectiveness and competitiveness of the proposed method against the compared classic subspace learning approaches.}
\end{abstract}

%
\IEEEpeerreviewmaketitle

\section{Introduction}
\IEEEPARstart{I}{n} recent years, the rapid development of the Internet has made multimedia data omnipresent in many social websites such as Facebook, YouTube and so on. We have witnessed the unprecedented growth of multi-modal data \cite{luo2017image,li2011no,sang2014no,luo2016novel,zheng2006reformative,shen2013content,li2013no,chen2018new,sun2011quantum,gao2004ant,zheng2006nearest}. In multimedia domain, numerous types of media data such as video clips, images, texts and so on, make up the multi-modal data. Representative image retrieval methods~\cite{Comaniciu2002Real,shu2011novel,zhang2022robust,zhang2021label} are not directly applied in multiple modality. Though multi-modal data have different forms, they are referred to the same topic or event. Therefore, various efforts have been taken to research cross-modal retrieval, which aims to achieve the required relevant objects from one modality when given one data object from another modality as query. The provided results can be helpful to the users to achieve useful events. In the past decades, cross-modal retrieval has attracted the attention of the researchers from both academia and industry. In this paper, we mainly focus on two typical cross-modal sub-retrieval tasks: image retrieves text(I2T) and text retrieves image(T2I).

There remains a fundamental problem for cross-modal retrieval approaches, i.e. how to make the similarity between multi-modal data be directly measured, which is referred to as the heterogeneity gap. To solve the problem, there comes to two strategies. The first traditional one is to calculate cross-modal similarity directly based on the known relationship by learning cross-modal approaches~\cite{jia2011learning,jiang2014relative,song2017multimodal}. The second learns a latent common subspace where the distance between multi-modal data can be executed, which is generally termed as cross-modal subspace learning. Generally speaking, according to whether semantic label is utilized, existing cross-modal subspace learning algorithms can be divided into two types: unsupervised approaches~\cite{akaho2006kernel,sharma2011bypassing,tenenbaum2000separating} and supervised approaches~\cite{wang2015joint,rasiwasia2010new,kang2015learning}. Specifically, the unsupervised approaches learn the intrinsic characteristic and correlation of multi-modal data. Even there exists the heterogeneity, some intrinsic correlations are still remained among multi-modal data because they express the same object. There are a series of outstanding techniques proposed to measure the dependence between multi-modal data, such as kullback Leibler(KL) divergence~\cite{amari1996new}, Hilbert-Schmidt Independence Criterion(HSIC)~\cite{davis2007information,principe2010information} and so on. In the following section, we will briefly introduce HSIC to measure the correlation between data from different modalities. Although unsupervised approaches have been effectively applied to cross-modal retrieval, their performance is not as pleasurable as expected due to the lack of available discriminative information. To find a solution to this problem, supervised approaches introduce semantic label information to study the discriminative feature of different modalities. Existing supervised approaches have shown the great effectiveness to cross-modal retrieval missions in the past few years. However, it is notable that even though the supervised information has been used in these approaches, many methods do not explore the semantically structural information and the correlation among multi-modal information during learning a latent subspace. The discriminative semantic information is not fully exploited as well. Actually, the semantically structural information is much vital to make the representation more discriminative in cross-modal learning methods.

\begin{figure*}[t]
\begin{center}
\includegraphics[trim={0mm 0mm 0mm 0mm},clip,width=1\linewidth]{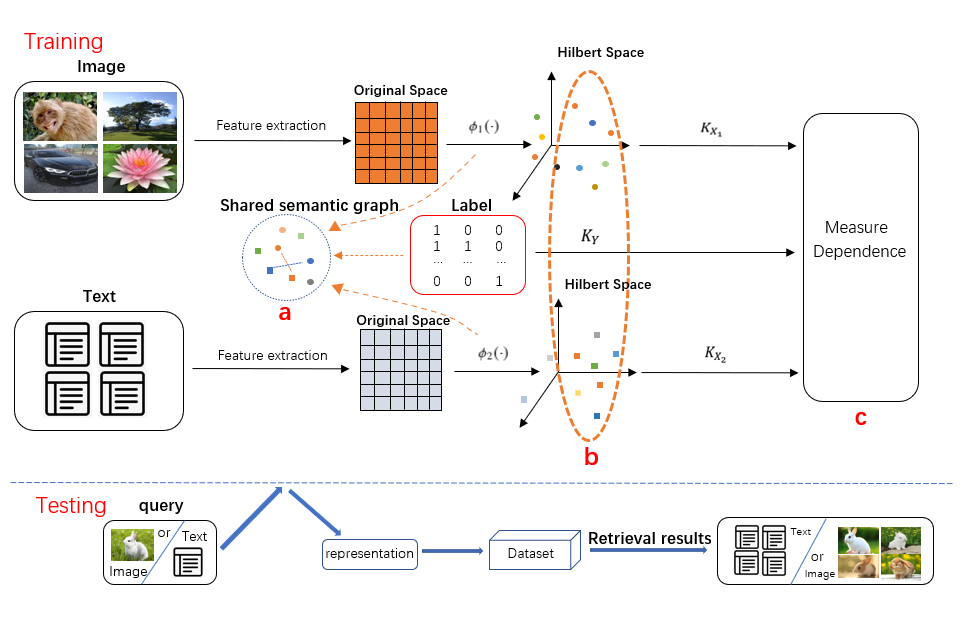}
\end{center}
\caption{Illustration of the framework of our DS²L. In the training phase, image modality and text modality are respectively mapped to the common Hilbert space. In the common Hilbert space, semantically structural information is preserved in each modality. (a) is a shared semantic graph constructed to learn the discriminative feature representation. (b) is a similarity preservation term to compensate for the shortcomings of insufficient use of discriminative data. (c) adopts HSIC to preserve the consistency of the similarity among samples for each modality. In the testing phase, taking data in one modality as a query set, we can achieve its feature representation and then the relevant data from another modality will be returned from the database.}\label{framework1}
\end{figure*}

To this end, in this paper we present such a novel framework called discriminative supervised subspace learning for cross-modal retrieval. It aims to not only preserve the correlation among multi-modal information but also fully exploit the semantically structural information. Specifically, to achieve this goal, it maximizes the inter-modality correlation and intra-modality similarity relationship to preserve the consistence between feature-similarity and semantic-similarity of samples. Furthermore, it makes the learned subspace representation preserve the semantic structure well. Following this learning strategy, the correlation among multi-modal data and the semantically structural information are fully exploited to ensure the learned representation to be discriminative. Fig.~\ref{framework1} depicts the framework of the proposed method. Given an arbitrary query, we can obtain its feature representation, and the relevant data of another type that is similar content to the query is returned from the database.
The main innovations of our proposed method can be summarised as follows:
\begin{itemize}
\item A discriminative supervised subspace learning architecture is proposed to joint feature selection and semantic structure preservation into subspace learning. A similarity preservation term is constructed to the learning model to learn the feature representation of each modality. Actually, this term can compensate for the shortcomings of insufficient use of discriminative data and make the learned representation more discriminative.  

\item An iterative optimization method based on the Stiefel manifold is designed for the optimization problem. It has an excellent convergence behaviour and we theoretically provide its rigorous convergence analysis.

\item Experiments are performed extensively on three publicly available benchmark datasets. The results demonstrate that our method is competitive against the compared classic subspace learning approaches, which indicates the effectiveness and superiority of the proposed method. 
\end{itemize}

The remainder of the paper is organised as follows:
In Section~\ref{relatedwork}, we introduce the related work in the field of cross-modal retrieval. 
Detailed mathematical formulations of our model and an optimization algorithm of the proposed DS²L method are presented in Section~\ref{method}.
The experimental results and their analysis are provided in Section~\ref{experiment}.
Last, the conclusions are drawn in Section~\ref{conclusion}.

\section{Related Work}
\label{relatedwork}
In the past decades, cross-modal retrieval has received great attention in literature due to its widespread use in practice. Various of methods have been studied for the retrieval tasks. Cross-modal retrieval approaches are roughly divided into two categories. One is hash code representation learning and the other is real-valued representation one. Thereinto, hash code representation learning based methods, also called as cross-modal hashing, are designed to learn a common Hamming space where the distance among multiple modalities can be measured. Semantics-preserving hashing(SePH)~\cite{lin2015semantics} is a representative method which transforms the affinity matrix into a probability distribution. Then the distribution is preserved in Hamming space via minimizing their Kullback-Leibler divergence. Supervised matrix factorization hashing(SMFH)~\cite{liu2016supervised} integrates the graph regularization and matrix factorization into a unified framework. These cross-modal hashing methods are more geared toward computational efficiency, thus they pay less attention to effectiveness. Since the representations are encoded to binary codes, the retrieval accuracy generally decreases due to the loss of information.

The proposed approach in this paper falls in the branch of real-valued representation learning, which is also termed as cross-modal subspace learning. To bridge the heterogeneous gap, real-valued representation learning based methods are proposed to find a common subspace where the different data among different modalities can be measured directly. Generally speaking, this type of methods can be divided into two branches: unsupervised subspace learning and supervised subspace learning. The unsupervised approaches only use co-occurrence information to learn common representations for different types of data. In~\cite{hardoon2004canonical}, canonical correlation analysis(CCA) studies the correlation coefficients to maximize the feature correlation among different modalities. In~\cite{lai2000kernel}, KCCA incorporates a kernel mapping based on CCA. It exploits a nonlinear mapping to map the original feature into a common subspace, and then establishes the modality correlations. Besides, Partial Least Squares(PLS)~\cite{sharma2011bypassing} linearly maps heterogeneous into a shared space where the directions of maximum covariance are found. Although these unsupervised approaches can obtain certain success, the discriminative explicit high-level semantics can not be captured due to the lack of discriminative ability. To solve this problem, supervised cross-modal retrieval approaches are developed to learn a more discriminative shared space with supervised semantic label information. The supervised methods enforce different-category samples to be transformed far apart while the same-category samples lie as close as possible. An example is~\cite{sharma2012generalized}, which proposed a generalized multiview analysis(GMA) and is a supervised extension of CCA. Since the real world is usually remarked by multiple labels, Multi-label CCA(ml-CCA)~\cite{ranjan2015multi} learns a common subspace with multi-label annotations. Semantic Correlation Matching(SCM)~\cite{pereira2013role} exploits logistic regression as the classification strategy to learn semantic representation. Moreover, in~\cite{yu2020cross}, cross-modal subspace learning via kernel correlation maximization and discriminative structure preserving(CKD) maximizes the correlation among multi-modal data and preserves the semantic information within each modality simultaneously. It preserves both the shared semantic structure and the correlation among multiple modalities.

Beyond that, due to Deep neural network(DNN), many deep learning methods have been developed to perform cross-modal retrieval in the past few years. As a non-linear extension of CCA, by incorporating DNN into CCA, deep canonical correlation analysis(DCCA)~\cite{andrew2013deep} is proposed to learn the complex non-linear transformations for each modality. In\cite{jiang2017deep}, Deep Cross-modal hashing(DCMH) is an end-to-end learning framework with DNN to integrate feature learning and hash-code learning into the same framework. Besides, Deep Supervised Cross-Modal Retrieval(DSCMR)~\cite{zhen2019deep} minimises the discrimination loss in both the label space and the common representation space to supervise the model learning discriminative features. Though such type of methods have shown their remarkable abilities, they are often confronted with the difficulty of high computational complexity because tuning parameters involved in neural network requires a lot of efforts during the training phase.

Unlike many of the existing subspace learning approaches, the supervised subspace learning method proposed in this paper is dedicated to compensate for the shortcomings of insufficient use of discriminative information and further make the learned representation more discriminative. Our model not only explores the correlation among multiple modalities, but also makes the learned subspace presentation well preserve the semantic structure.

\section{The Proposed [DS²L] Method}
\label{method}

\subsection{Problem Formulation}
We make a assumption that there are m modalities, donated as  $\mathrm{X}=\left[X^{(1)},X^{(2)},\ldots,X^{(m)}\right]$. The $v$-th modality modality $\mathrm{X}^{(v)}=\left[x_1^{(v)},x_2^{(v)},\ldots,x_n^{(v)}\right]\in \mathbb{R}^{n \times d_{v}}(v=1, \ldots, m)$ contains n samples with $d_{v}$ dimension. The shared semantic label matrix is represented by $\mathrm{Y}=[y_1,y_2,\ldots,y_n]^\top\in \mathbb{R}^{n \times c}$, where c is the total number of semantic categories. If the $j$-th instance belongs to the $k$-th semantic category, we have $y_{jk}=1$, otherwise we have $y_{jk}=0$. The target of our paper is to learn isometric representation for heterogeneous data of multiple modalities. Without loss of generality, the samples of each modality are zero-centered, i.e., $\sum_{i=1}^{n} X_{i}^{(v)}=0(v=1,2 \ldots N)$. 

\subsection{Hilbert-Schmidt independence criteria(HSIC)}
Let $\phi(y)$ and $\phi(z)$ be two mapping functions with $\phi(y):y\rightarrow \mathbb{R}^{d}$ and $\phi(z):z\rightarrow \mathbb{R}^{d}$. The associated positive definite kernels $K_y$ and $K_z$ are respectively defined as $\mathrm{K}_y=<\phi(y),\phi(y)>$ and $\mathrm{K}_z=<\phi(z),\phi(z)>$. Let $C_{yz}$ be the cross-covariance function between $y$ and $z$.  $C_{yz}$ is defined as $C_{yz}=\mathbb{E}_{yz}[(\phi(y)-u_y)\otimes(\phi(z)-u_z)]$, where $u_y$ and $u_z$ are the expectation of $\phi(y)$ and $\phi(z)$ respectively, i.e., $u_y=\mathbb{E}(\phi(y))$ and $u_z=\mathbb{E}(\phi(z))$. Besides, $\otimes$ represents the tensor product. Then we can define HSIC, which is the Hilbert-Schdmidt norm of $C_{yz}$, as: $\mathrm{HSIC}=\|C_{y z}\|_{H S}^{2}$. For any matrix $A$, we have $\|\mathbf{A}\|_{H S}=\sqrt{\sum_{i, j} a_{i j}^{2}}$. In practical applications, an empirical estimate formulation of HSIC is exploited very often. Given $n$ paired data samples $\mathrm{D}=\{(y_1,z_1),\ldots,(y_n,z_n)\}$, the empirical expression of HSIC is formulated as:
\begin{equation}
\textbf{HSIC}=(n-1)^{-2}\mathrm{tr}(K_y H K_z H)
\end{equation}
where $K_y$ and $K_z$ are two Gram matrices, $\mathrm{H}=\boldsymbol{I}-\frac{1}{n} \mathbf{1}_{n} \mathbf{1}_{n}^{T}$ is a centering matrix, and $\mathbf{1}_{n}\in \mathbb{R}^n$ is a full-one column vector. More details of HSIC can refer to~\cite{davis2007information,principe2010information}.

\subsection{Model}
For simplicity, in the following part we focus on our algorithm for bimodal data, specifically for image and text.
\subsubsection{Kernel dependence Maximization}
The similarity between two kernel matrices can be calculated with measures, e.g., kernel alignment, Euclidean distance, Kullback-Leibler divergence and so on. In this paper, we use the Hilbert-Schmidt independence criteria(HSIC) which is briefly introduced in the preceding subsection to maximize the kernel correlation among multi-modal data. Multi-modal data are projected into the Hilbert space under the mapping functions $\phi_v(\cdot)(v=1,2)$. $\phi_1(\cdot)$ is constructed for image modality and $\phi_2(\cdot)$ for text modality. For simplicity, we adopt linear kernel as the kernel measure. Then the kernel matrix can be defined as $\mathrm{K}_{x_v}=<W_v,W_v>=W_v W_v^{T}$,where $\mathrm{W}_v=\phi_v(X^{(v)})=X^{(v)}P_v(v=1,2)$ and $P_v$ are the projection matrices. Similar to the previous formulations, the kernel matrix of the semantic label can be denoted as $\mathrm{K}_Y=<Y,Y>=YY^{T}$. As a kernel itself in essence represents a similarity function, the similarity relationship among samples of each modality is measured by the kernel matrix, which is also termed as the intra-modality similarity. Our proposed method exploit the maximization of kernel dependence to preserve the intra-modality similarity relationship. According to Eq.1, the objective formulation can be signified as:
\begin{equation}
\begin{aligned}
&\max _{P_{1}, P_{2}}\operatorname{tr}(H K_{X_{1}} H K_{X_{2}})+\operatorname{tr}(H K_{X_{1}} H K_{Y})+\operatorname{tr}(H K_{X_{2}} H K_{Y})\\
&\textrm{s.t.}\ \ P_1^{T}P_1=\boldsymbol{I},P_2^{T}P_2=\boldsymbol{I}
\end{aligned}
\end{equation}
where $\mathrm{K}_{X1}=X_1P_1P_1^{T}X_1^{T}$, $\mathrm{K}_{X2}=X_2P_2P_2^{T}X_2^{T}$, $\mathrm{K}_{Y}=YY^{T}$. Specifically, the first term is to measure the correlation between multi-modal data. The second and the third term is to measure the correlation between each modality and the shared semantic label. By using Eq.2, we can take the inter-modality similarity and the intra-modality similarity into consideration simultaneously.

\subsubsection{Discriminative Structure Preservation}
It is worth noting that different modalities share the same semantic information in spite of locating isomeric spaces. Considering the fact that the semantically structural information plays an important role in subspace representation learning, we hope that we can well preserve the semantically structural relationship in Hilbert space among samples. In other words, when the semantic relationship among samples is closer, their distances are supposed to be nearer in the shared Hilbert space. To achieve this goal, first we employ the semantic label vectors among samples to calculate their cosine similarity. Let $y_m$ be the $m$-th sample and $y_n$ be the $n$-th sample, we have such a formulation to measure the similarity between $y_m$ and $y_n$:
\begin{equation}
\mathrm{S}_{y_m,y_n}=\frac{y_{m} \cdot y_{n}}{\|y_{m}\|_{2}\|y_{n}\|_{2}}
\end{equation}
where $\|y_{m}\|_{2}$ represent the $L_2$-norm of the vector $y_m$. Then the objective function can be formulated via Eq.3 as follows:
\begin{equation}
\begin{aligned}
&\min _{P_{1}, P_{2}}\alpha_1\operatorname{tr}(P_1^{T}X_1^{T} L X_1P_1)
+\alpha_2\operatorname{tr}(P_2^{T}X_2^{T} L X_2P_2)\\
&\textrm{s.t.}\ \ P_1^{T}P_1=\boldsymbol{I},P_2^{T}P_2=\boldsymbol{I}
\end{aligned}
\end{equation}
where $L=\operatorname{diag}(S\mathbf{1})-S$ represents a graph Laplacian matrix and $\alpha_1$ and $\alpha_2$ are two adjustable parameters. According to some discussions in~\cite{song2017semi,zhang2020unified}, models based on the $l_{2,1}$-norm have demonstrated their effectiveness of sparsity, feature selection and robustness to noise. Therefore, in this paper we impose the $l_{2,1}$-norm constraint on the projection matrices of our model, so that the model can be expected to learn more discriminative subspace representation by removing the redundant features. With the addition of the $l_{2,1}$-form constraint,  Eq.4 can be extended to the following formulation:
\begin{equation}
\begin{aligned}
\min _{P_{1}, P_{2}}&\alpha_1(\operatorname{tr}(P_1^{T}X_1^{T} L X_1P_1)+\lambda_1\|P_{1}\|_{2,1})\\
&+\alpha_2(\operatorname{tr}(P_2^{T}X_2^{T} L X_2P_2)+\lambda_2\|P_{2}\|_{2,1})\\
&\textrm{s.t.}\ \ P_1^{T}P_1=\boldsymbol{I},P_2^{T}P_2=\boldsymbol{I}
\end{aligned}
\end{equation}
where $\lambda_1$ and $\lambda_2$ are the regularization parameters.

\subsubsection{Similarity Preservation}
CKD~\cite{yu2020cross} has proposed a learning model which combines Eq.2 and Eq.5 into a joint framework, and have achieved certain improvement on subspace learning. However, CKD does not obtain satisfactory results in the further study because of the insufficient use of supervised information. To overcome this shortcoming and improve the discrimination of the learned representation for cross-modal retrieval,  the problem can be transformed to the following function:
\begin{equation}
\begin{aligned}
&\min _{P_{1}, P_{2}}\theta\|YY^{T}-X_{1}P_{1}(X_2P_2)^{T}\|_{F}^{2}\\
&\textrm{s.t.}\ \ P_1^{T}P_1=\boldsymbol{I},P_2^{T}P_2=\boldsymbol{I}
\end{aligned}
\end{equation}
where $\theta$ is a trade-off parameter.

\subsubsection{Overall Objection Function}
By combining Eq.2, Eq.5 and Eq.6, we can integrate kernel dependence maximization, discriminative structure preservation and similarity preservation into a joint framework. Therefore, the supervised information and the semantic structure are fully exploited to learn a discriminative common subspace which preserves both the shared semantic structure and the dependence among multiple modalities. We formulate the overall objective function of our proposed DS²L as:
\begin{equation}
\begin{aligned}
\min _{P_{1}, P_{2}}&\alpha_1(\operatorname{tr}(P_1^{T}X_1^{T} L X_1P_1)+\lambda_1\|P_{1}\|_{2,1})\\
&+\alpha_2(\operatorname{tr}(P_2^{T}X_2^{T} L X_2P_2)+\lambda_2\|P_{2}\|_{2,1})\\
&+\theta\|YY^{T}-X_{1}P_{1}(X_2P_2)^{T}\|_{F}^{2}\\
&+\beta(-\operatorname{tr}(H K_{X_{1}} H K_{X_{2}})-\operatorname{tr}(H K_{X_{1}} H K_{Y})\\
&-\operatorname{tr}(H K_{X_{2}} H K_{Y}))\\
&\textrm{s.t.}\ \ P_1^{T}P_1=\boldsymbol{I},P_2^{T}P_2=\boldsymbol{I}
\end{aligned}
\end{equation}
where $\beta$ is an adjustable parameter. For simplicity, we set $\beta=1$. To make the optimization introduced below performed more conveniently, by virtue of $\|W\|_{2,1}=\operatorname{tr}(W^{T}DW)$, where the intermediate variable $D=\operatorname{diag}(\frac{1}{\|W^{\cdot\cdot i}\|_{2}})$, we rewrite Eq.7 as:
\begin{equation}
\begin{aligned}
\min _{P_{1}, P_{2}}&-\operatorname{tr}(H K_{X_{1}} H K_{X_{2}})-\operatorname{tr}(H K_{X_{1}} H K_{Y})\\
&-\operatorname{tr}(H K_{X_{2}} H K_{Y}))\\
&+\alpha_1(\operatorname{tr}(P_1^{T}X_1^{T} L X_1P_1)+\lambda_1\operatorname{tr}(P_1^{T}D_1 P_1))\\
&+\alpha_2(\operatorname{tr}(P_2^{T}X_2^{T} L X_2P_2)+\lambda_2\operatorname{tr}(P_2^{T}D_2 P_2))\\
&+\theta\|YY^{T}-X_{1}P_{1}(X_2P_2)^{T}\|_{F}^{2}\\
&\textrm{s.t.}\ \ P_1^{T}P_1=\boldsymbol{I},P_2^{T}P_2=\boldsymbol{I}
\end{aligned}
\end{equation}
where $D_{v}=\operatorname{diag}(\frac{1}{\|P_{v}^{\cdot\cdot i}\|_{2}})$ and $P_{v}^{\cdot\cdot i}$ is used to represent its $i$-th row$(v=1,2)$.

\subsection{Optimization}
According to Eq.8, we can summarize the objective function as such a form:
\begin{equation}
\begin{aligned}
&\min _{W}f(W)\\
&\textrm{s.t.}\ \ W^{T}W=\boldsymbol{I}_{m}
\end{aligned}
\end{equation}
where $f(W)$ is the objective function and $W$ are the parameters to be optimized. In Euclidean space, the problems of the form of Eq.9 are typically cast as eigenvalue problems. However, according to the literatures~\cite{gao2019robust,harandi2017dimensionality}, we know that one of the optimization methods formulates problems with unitary constraints as optimization problems on the Stiefel manifold. We naturally apply the conjugate gradient method on the Stiefel manifold for optimization in this paper. The Stiefel manifold is formed by a set of all $n\times r$ semi-orthogonal matrices, which can be denoted as: $St(n,r)=\{W\in\mathbb{R}^{n\times m}: W^{T}W=\boldsymbol{I}_{m}\}$. Thanks to the matlab toolbox named manopt~\cite{boumal2014manopt}, given the cost function and the derivatives, our optimization of this problem can be easily and efficiently implemented.

Since there are two parameters to be optimized, we employ the alternative optimization strategy to optimize the objective function in Eq.8. According to the alternative optimization rules, each time we learn one parameter with the other parameters fixed. Specifically, we decompose the original optimization problem into the following two sub-minimization problems: solve $P1$ with $P2$ fixed and solve $P2$ with $P1$ fixed.

\subsubsection{Optimization of $P1$ with $P2$ fixed}
Keeping only the terms relating to $P1$, we can achieve the following equivalent optimization problem of Eq.8:
\begin{equation}
\begin{aligned}
&\min(\theta\|YY^{T}-X_1P_1P_2^{T}X_2^{T}\|_F^{2}-\operatorname{tr}(P_1^{T}Q_1P_1))\\
&\textrm{s.t.}\ \ P_1^{T}P_1=\boldsymbol{I}
\end{aligned}
\end{equation}
where $Q_1=X_1^{T}HX_2P_2P_2^{T}X_2^{T}HX_1+X_1^{T}HYY^{T}HX_1-\alpha_1X_1^{T}LX_1-\alpha_1\lambda_1D_1$.

For simplicity, we let $\mathcal{L}_1=\theta\|YY^{T}-X_1P_1P_2^{T}X_2^{T}\|_F^{2}-\operatorname{tr}(P_1^{T}Q_1P_1)$.
The Euclidean gradient of $\mathcal{L}_1$ with respect to $P_1$ can be computed as:
\begin{equation}
\begin{aligned}
\frac{\partial \mathcal{L}_1}{\partial \boldsymbol{P}_{1}}=-2X_1^{T}(YY^{T}-X_1P_1B^T)B-2Q_1P_1
\end{aligned}
\end{equation}
where $B=X_2P_2$.
For the conjugate gradient method on the Stiefel manifold used to optimize $P_1$, we briefly summarize its implementation into the following steps:

\begin{itemize}
\item Computing the Stiefel gradient $\tilde{\nabla} L(P_1^{k})$ at the $k$-th iteration:
\begin{equation}
\tilde{\nabla} L(P_1^{k})=\frac{\partial \mathcal{L}_1}{\partial \boldsymbol{P}_{1}^{k}}-P_1^{k}\frac{\partial \mathcal{L}_1}{\partial \boldsymbol{P}_{1}^{k}}^T P_1^{k}
\end{equation}

\item Using parallel transport to obtain the new search direction:
\begin{equation}
H_k=-\tilde{\nabla} L(P_1^{k})+\gamma_{k-1} \Gamma(H_{k-1},P_1^{k-1},P_1^{k})
\end{equation}
where $H_{k-1}$, $H_k$ respectively denote the previous search direction and the currently learned search direction, and $\Gamma(H_{k-1},P_1^{k-1},P_1^{k})$ represents the parallel transport.

\item Utilizing line search along the geodesic at $P_1^{k}$ in the direction $H_k$ to find $P_1^{k}=\mathop{\arg\min}\mathcal{L}_1P_1^{k}(t)$, where
\begin{equation}
P_1^{k}(t)=P_1^{k}M(t)+QN(t)
\end{equation}
and $M(t)$ and $N(t)$ are given by: 
\begin{equation}
\left(\begin{array}{c}M(t) \\ N(t)\end{array}\right)=\exp t\left(\begin{array}{cc}A & -R^{T} \\ R & 0\end{array}\right)\left(\begin{array}{c}I_{p} \\ 0\end{array}\right)
\end{equation}
where $A=(P_1^{k})^T H_{k-1}$.

\item Repeating the aforementioned steps until convergence to a local minimum, or reach to the maximum number of iterations.
\end{itemize}

\subsubsection{Optimization of $P2$ with $P1$ fixed}
Keeping only the terms relating to $P2$, we can achieve the following equivalent optimization problem of Eq.8:
\begin{equation}
\begin{aligned}
&\min(\theta\|YY^{T}-X_1P_1P_2^{T}X_2^{T}\|_F^{2}-\operatorname{tr}(P_2^{T}Q_2P_2))\\
&\textrm{s.t.}\ \ P_2^{T}P_2=\boldsymbol{I}
\end{aligned}
\end{equation}
where $Q_2=X_2^{T}HX_1P_1P_1^{T}X_1^{T}HX_2+X_2^{T}HYY^{T}HX_2-\alpha_2X_2^{T}LX_2-\alpha_2\lambda_2D_2$.

We let $\mathcal{L}_2=\theta\|YY^{T}-X_1P_1P_2^{T}X_2^{T}\|_F^{2}-\operatorname{tr}(P_2^{T}Q_2P_2)$.
The Euclidean gradient of $\mathcal{L}_2$ with respect to $P_2$ can be computed as:
\begin{equation}
\begin{aligned}
\frac{\partial \mathcal{L}_2}{\partial \boldsymbol{P}_{2}}=-2X_2^{T}(Y^{T}Y-X_2P_2B^T)B-2Q_2P_2
\end{aligned}
\end{equation}
where $B=X_2P_2$.
Due to the optimization strategy of $P_2$ is same as that of $P_1$ mentioned above, we will not go into the details here. 

To better understand the procedure of the optimization problem, we briefly outline the solver in Algorithm 1. 
\begin{algorithm}
\caption{The algorithm proposed in this paper(DS²L)}
\hspace*{0.02in}{\bf Input:}
The training data $X^{(v)} \in \mathbb{R}^{n \times d_v}$; the label matrix $Y \in \mathbb{R}^{n \times c}$; the demension of the common Hibert space d; parameters $\alpha_v$, $\lambda_v$ and $\theta$.($v=1,2$)\\
\hspace*{0.02in}{\bf Output:}
$P_1$,$P_2$
\begin{algorithmic}
\STATE Initialize  $P_1$,$P_2$, and $t=0$.\\
Calculating similarity matrix $S$ according to Eq.3.
\WHILE{the objective function not converged}
\STATE 1: Compute $D_1$ by $D_1^{ii}=\frac{1}{2\left\|P_{1}^{\cdots i}\right\|_{2}}$.
\STATE 2: Compute $D_2$ by $D_2^{ii}=\frac{1}{2\left\|P_{2}^{\cdots i}\right\|_{2}}$.
\STATE 3: Update P1: calculate the gradient with Eq.11.\\
\STATE 4: Obtain $P_1^{(t+1)}$ by using conjugate gradient method on the Stiefel manifold.
\STATE 5: Update P2: calculate the gradient with Eq.17.\\
\STATE 6: Obtain $P_2^{(t+1)}$ by using conjugate gradient method on the Stiefel manifold.
\STATE 7: $t=t+1$.
\ENDWHILE
\STATE Return $P_1$, $P_2$.
\end{algorithmic}
\end{algorithm}
\subsection{Convergence analysis}
The detailed optimization procedure is shown in Algorithm 1. The process is repeated until the algorithm converges. The convergence can be summarized by the following Theorem 1.

\newtheorem{thm}{\bf Theorem}
\begin{thm}\label{thm1}
The objection function defined by Eq.8 based on the iterative optimizing rules in Algorithm 1 is decreasing monotonically, and it will converge to the global minimum.
\end{thm}

For the proof of Theorem 1, we introduce such a lemma which is demonstrated in~\cite{nie2010efficient}. 

\newtheorem{lemma}{\bf Lemma}
\begin{lemma} \label{lemma1}
For any nonzero $a\in \mathbb{R}^{t}$ and $b\in \mathbb{R}^{t}$, we have
\begin{equation}
\begin{aligned}
\frac{\|a\|_{2}}{2\|b\|_{2}}-\|a\|_{2} \geq \frac{\|b\|_{2}}{2\|b\|_{2}}-\|b\|_{2}
\end{aligned}
\end{equation}
\end{lemma}
With the introduction of the lemma above, we can prove Theorem 1.

\begin{proof}
As shown in the former subsection, the optimization of $P_1$ and $P_2$ in Algorithm 1 are symmetrical. Therefore, only one of them needs to be proved. Below we will prove the optimization of $P_1$ in detail. 

By virtue of Eq.10, we  can let $\mathrm{G}(P_1)=\operatorname{tr}(HK_{X_1}HK_{X_2})+\operatorname{tr}(HK_{X_1}HK_{Y})-\alpha_1\operatorname{tr}(P_1^{T}X_1^{T}LX_1P_1)-\theta\|YY^{T}-X_1P_1P_2^{T}X_2^{T}\|_F^{2}$. Then the objection function turns to be $\mathrm{H}(P_1)=\lambda_1\operatorname{tr}(P_1^{T}D_1P_1)-G(P_1)$. For the $m$-th iteration, there goes:
\begin{equation}
\begin{aligned}
P_{1}^{(m+1)} &=\underset{P_{1}}{\arg \min }(\lambda_{1} \operatorname{tr}(P_{1}^{T} D_{1} P_{1})-G(P_{1})) \\ & \Rightarrow \lambda_{1} \operatorname{tr}(P_{1}^{(m+1)^{T}} D_{1}^{(m)} P_{1}^{(m+1)})-G(P_{1}^{(m+1)}) \\ & \leq \lambda_{1} \operatorname{tr}(P_{1}^{(m)^{T}} D_{1}^{(m)} P_{1}^{(m)})-G(P_{1}^{(m)}) \\ & \Rightarrow \lambda_{1} \sum_{i}^{d_{1}} \frac{\|P_{1}^{\cdot \cdot i(m+1)}\|_{2}^{2}}{2\|P_{1}^{\cdot \cdot i(m)}\|_{2}^{2}}-G(P_{1}^{(m+1)}) \\ & \leq \lambda_{1} \sum_{i}^{d_{1}} \frac{\|P_{1}^{\cdot \cdot i(m)}\|_{2}^{2}}{2\|P_{1}^{\cdot \cdot i(m)}\|_{2}^{2}}-G(P_{1}^{(m)}) \\ & \Rightarrow \lambda_{1}(\sum_{i}^{d_{1}} \frac{\|P_{1}^{\cdot \cdot i(m+1)}\|_{2}^{2}}{2\|P_{1}^{\cdot \cdot i(m)}\|_{2}^{2}}-\|P_{1}^{\cdot \cdot i(m+1)}\|_{2,1}) \\ &+\lambda_{1}\|P_{1}^{\cdot \cdot i(m+1)}\|_{2,1}-G(P_{1}^{(m+1)}) \\ & \leq \lambda_{1}(\sum_{i}^{d_{1}} \frac{\|P_{1}^{\cdot \cdot i(m)}\|_{2}^{2}}{2\|P_{1}^{\cdot \cdot i(m)}\|_{2}^{2}}-\|P_{1}^{\cdot \cdot i(m)}\|_{2,1}) \\ &+\lambda_{1}\|P_{1}^{\cdot\cdot i(m)}\|_{2,1}-G(P_{1}^{(m)})
\end{aligned}
\end{equation}
According to Lemma 1, the following formulation can be obtained:
\begin{equation}
\begin{aligned}
&\lambda_{1}\|P_{1}^{\cdot \cdot i(m+1)}\|_{2,1}-G(P_{1}^{(m+1)}) \leq \lambda_{1}\|P_{1}^{\cdot \cdot i(m)}\|_{2,1}-G(P_{1}^{(m)})
\end{aligned}
\end{equation}
then we have
\begin{equation}
H(P_1^{(m+1)})\leq H(P_1^{(m)})
\end{equation}
Likewise, when updating $P_2$ with $P_1$ fixed, we can prove that $H(P_2^{(m+1)})\leq H(P_2^{(m)})$ as well. Further, we have $ H(P_1^{(m+1)},P_2^{(m+1)})\leq H(P_1^{(m)},P_2^{(m)})$.
\end{proof}

In summary, the overall objective function based on Algorithm1 is non-increasing. Considering that the optimization problem is convex, we can complete the proof that the objective function will converge to its global optimal solution in the end.




\section{Experiments}
\label{experiment}
In this section, experiments are conducted on three benchmark datasets, i.e., NUS-WIDE~\cite{chua2009nus}, MIRFlickr~\cite{huiskes2008mir} and Pascal-Sentence~\cite{wei2016cross}, to test the performance of our DS²L for cross-modal retrieval. We introduce two widely-used evaluation protocols to evaluate the algorithm. Given a problem, with the use of Algorithm 1, we can learn the projection matrices on the training set. Then data from different modalities can be projected into a common subspace, where  the correlation of projected data from each modality can be measured. In the testing phase, taking data in one modality as a query set, we can retrieve the relevant data from another modality. To measure the similarity among data, in this paper we adopt the normalized correlation (NC), which shows promising performance for cross-modal retrieval as demonstrated in~\cite{pereira2013role}.

\begin{table}
\footnotesize
\renewcommand{\arraystretch}{1.2}
\caption{Statics of datasets.}
\label{statics}
\centering
\begin{tabular}{lcccccc}
\hline
Datasets  & d1 & d2 & \#labels & \#total & \#training & \#testing \\
\hline
NUS-WIDE & 500 & 1000 & 21 & 8687 & 5212 & 3475 \\
MIRFlickr & 150 & 500 & 24 & 16738 & 5000 & 836 \\
Pascal-Sentence & 4096 & 100 &20 & 1000 & 600 & 400 \\
\hline
\end{tabular}
\end{table}

\subsection{Datasets}
This subsection will describe the three benchmark datasets we use briefly. Table $\uppercase\expandafter{\romannumeral1}$ plots the statistical information of these datasets.

\textbf{NUS-WIDE}: This is a subset from a real-world web image databases. It includes 190,420 image examples in total and each with 21 possible labels. For each pair, 500-dimensional SIFT BoVW features are extracted for image and 1000-dimensional text annotations for text. For simplicity, here we sample a subset with 8,687 image-text pairs which are further divided into two sections: 5,212 pairs for training and 3,475 pairs for testing.

\textbf{MIRFlickr}: This dataset consists of 25,000 images captured from the Flickr website. Each image and its corresponding textual tag constitute an image-text pair. Each pair is classified into some of the 24 classes. Only those pairs whose textual tags appear at least 20 times will be preserved. For each pair, image is represented by a 150-dimensional edge histogram vector and text is represented as a 500-dimensional vector obtained from PCA on the bag-of-words vectors. 5\% of the instances are chosen as the query set and 30\% as the training set.

\textbf{Pascal-Sentence}: This dataset has 1,000 image-text pairs. It consists of 20 semantic categories and each categories is comprised of  50 samples. From each semantic class we randomly select 30 pairs as training sets, and the remaining 20 pairs as testing sets. A convolutional neural network is employed to extract 4096-dimension CNN features to represent each image. Each text is represented with 100-dimensional LDA feature.


\subsection{Evaluation metric}
In the area of information retrieval, Mean Average Precision (MAP) and Cumulative Match Characteristic Curve(CMC) are two of the commonly used evaluation metrics. In our experiments, we adopt these metrics  to evaluate the performance of the retrieval. 

Given a set of queries, MAP refers to the average precision (AP) of all queries. Specifically, the AP of M returned results is defined as:
\begin{equation}
\begin{aligned}
AP=\frac{1}{R} \sum_{j=1}^{M} p(j) \delta(j)
\end{aligned}
\end{equation}
where $R$ represents the number of the corresponding samples in the retrieval set. $p(j)$ is the precision of the top $j$ returned query items. If the $j$-th  returned item is relevant to the query, we have $\delta(j)=1$, otherwise $\delta(j)$ is set to be $0$. The MAP score is calculated by averaging AP scores of all the queries.

CMC is the probability statistics that the true retrieval results appear in candidate lists of different sizes. Specifically, given query data, we suppose that this query can match true object if the retrieval result contains one or more objects classified into the same class. Assuming that the length of retrieval results is fixed as l, the rate of true match in all query is denoted as $CMC_{rank-l}$.

\begin{table}
\footnotesize
\renewcommand{\arraystretch}{1.2}
\caption{Summarization of compared methods.}
\label{summarize}
\centering
\begin{tabular}{lcccc}
\hline
Methods  & Unsupervised & Supervised & Kernel-based & Correlation-based \\
\hline
\textbf{CCA} & Y & N & N & Y \\
\textbf{KCCA} & Y & N & Y & Y \\
\textbf{ml-CCA} & N & Y & N & Y \\
\textbf{KDM} & N & Y & Y & Y \\
\textbf{CKD} & N & Y & Y & Y \\
\textbf{DS²L} & N & Y & Y & Y \\
\hline
\end{tabular}
\end{table}

\subsection{Experimental setup}
Our proposed DS²L in this paper is supervised, kernel-based and correlation-based. The methods used to compare with our approach are listed as follows: CCA~\cite{hardoon2004canonical}, KCCA~\cite{lai2000kernel}, ml-CCA~\cite{ranjan2015multi}, KDM~\cite{xu2018subspace} and CKD~\cite{yu2020cross}. Table \uppercase\expandafter{\romannumeral2} makes a summary about all the compared methods, where ‘Y’ is YES, indicating one method belongs to the corresponding ascription and ‘N’ is NO, indicating one method does not belong to the corresponding ascription. We also compare our approach with the case where $\beta$ is set to be $0$, and the case where we let $\alpha_1=\alpha_2=0$. We tune the parameter d which represents the dimension of the common space from the range of $\{10,20,30,40,50,60\}$. $\lambda_1$ and $\lambda_2$ are two coefficients of the regularization term in Eq.5 and in the experiment we fix them to $0.01$. $\alpha_1$ and $\alpha_2$ are two adjustable parameters in Eq.4. Empirically, we determine $\alpha_1=1$ and $\alpha_2=10$, as well as $\alpha_1=10^{-5}$ and $\alpha_2=1$ on NUS-WIDE and MIRFlickr, respectively. $\theta$ is a trade-off parameter in the objective function. We tune $\theta$ from $\{10^{-5},10^{-4},10^{-3},10^{-2},10^{-1},1,10,10^{2},10^{3},10^{4},10^{5}\}$. In the following part, we will show the best results of the experiments and give the parameter sensitivity analysis on $\alpha_1, \alpha_2, \theta$ and d. The experiments are performed on MATLAB 2020a and Windows 10 (64-Bit) platform based on desktop machine with 16 GB memory
and 4-core 2.9GHz CPU, and the model of the CPU is Intel(R) Core(TM) i5-9400.

\begin{table}
\footnotesize
\renewcommand{\arraystretch}{1.2}
\caption{The results of MAP comparison on NUS-WIDE and MIRFlickr.}
\label{mapcom}
\centering
\begin{tabular}{|l|c|c|c|c|}
\hline
Datasets                 & Methods & I2T & T2I & Average \\
\hline
\multirow{6}{*}{NUS-WIDE} & CCA     & 0.3099   & 0.3103   & 0.3101    \\
                          & KCCA    & 0.3088   & 0.3174   & 0.3096    \\
                          & ml-CCA  & 0.2787   & 0.2801   & 0.2794    \\
                          & KDM     & 0.3247   & 0.3118   & 0.3183    \\
                          & CKD     & 0.4149   & 0.4211   & 0.4180    \\
                          & DS²L  & \textbf{0.4488}   & \textbf{0.4514}   & \textbf{0.4501}   \\
\hline
\multirow{6}{*}{MIRFlickr} & CCA     & 0.5466   & 0.5477   & 0.5472    \\
                          & KCCA    & 0.5521   & 0.5529   & 0.5525    \\
                          & ml-CCA  & 0.5309   & 0.5302   & 0.5306    \\
                          & KDM     & 0.5951   & 0.5823   & 0.5887    \\
                          & CKD     & 0.6103   & 0.5933   & 0.6018    \\
                          & DS²L  & \textbf{0.6288}   & \textbf{0.6093}   & \textbf{0.6191}   \\
\hline
\end{tabular}
\end{table}

\begin{table}
\footnotesize
\renewcommand{\arraystretch}{1.2}
\caption{The results of the ablation experiments on the terms in Eq.7.}
\label{ablacom}
\centering
\begin{tabular}{|l|c|c|c|c|}
\hline
Datasets                 & Methods & I2T & T2I & Average \\
\hline
\multirow{4}{*}{NUS-WIDE} & DS²L($\alpha_1=\alpha_2=0$) &0.3813 &0.3802 &0.3808   \\
                          & DS²L($\beta=0$) &0.3904 &0.3993 &0.3948   \\
                          & DS²L($\theta=0$) &0.4149 &0.4211 &0.4180   \\
                          & DS²L  & \textbf{0.4488}   & \textbf{0.4514}   & \textbf{0.4501}   \\
\hline
\multirow{4}{*}{MIRFlickr} & DS²L($\alpha_1=\alpha_2=0$) &0.6018 &0.5921 &0.5971   \\
                          & DS²L($\beta=0$) &0.5981 &0.5992 &0.5986   \\
                          & DS²L($\theta=0$) &0.6103 &0.5933 &0.6018   \\
                          & DS²L  & \textbf{0.6288}   & \textbf{0.6093}   & \textbf{0.6191}   \\
\hline
\end{tabular}
\end{table}

\begin{figure*}[!t]
\begin{center}
\includegraphics[trim={0mm 0mm 0mm 0mm},clip,width=1\linewidth]{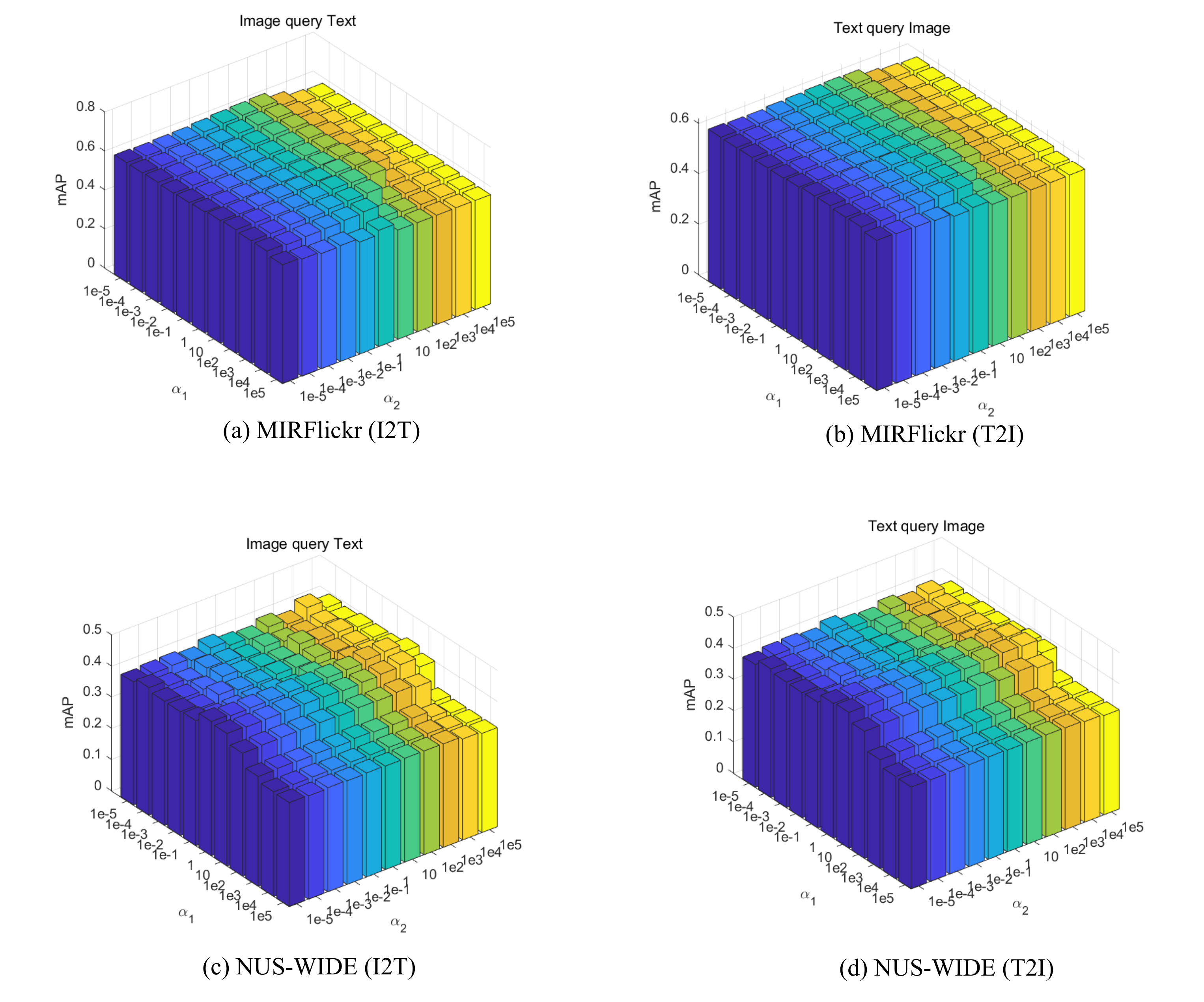}
\end{center}
\caption{Performance variation of DS²L with respect to $\alpha_1$ and $\alpha_2$ on three benchmark datasets.(a)I2T on Pascal-Sentence. (b)I2T on MIRFlickr. (c)I2T on NUS-WIDE. (d)T2I on Pascal-Sentence. (e) T2I on MIRFlickr. (f)T2I on NUS-WIDE.}\label{paracom1}
\end{figure*}

\begin{figure*}[!t]
\begin{center}
\includegraphics[trim={0mm 0mm 0mm 0mm},clip,width=1\linewidth]{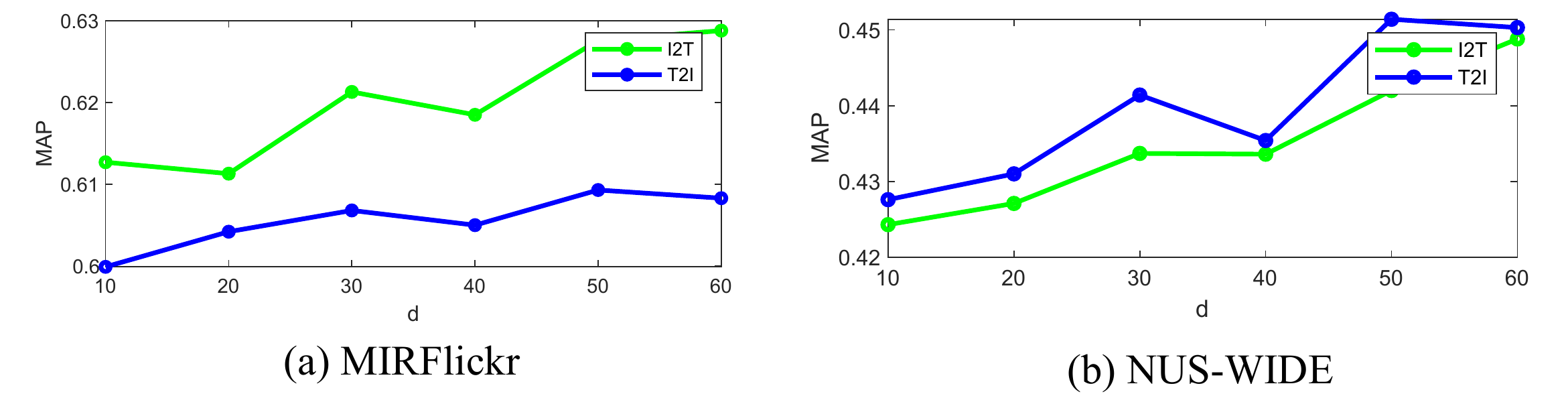}
\end{center}
\caption{Performance variation of DS²L with respect to $d$ on three benchmark datasets. (a)Pascal-Sentence, (b)MIRFlickr, (c)NUS-WIDE.}\label{parad}
\end{figure*}

\begin{figure*}[!t]
\begin{center}
\includegraphics[trim={0mm 0mm 0mm 0mm},clip,width=1\linewidth]{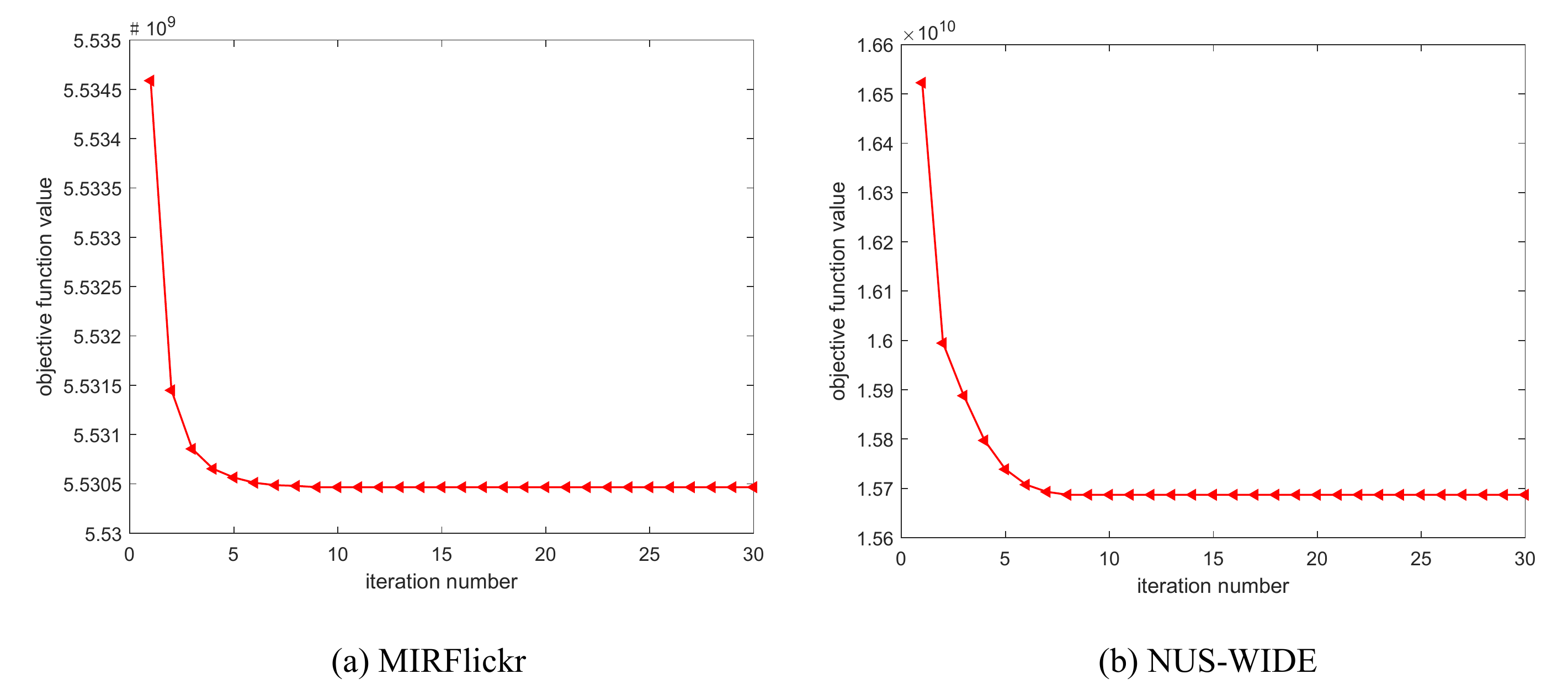}
\end{center}
\caption{The convergence of Algorithm 1 on three benchmark datasets, i.e., \textbf{NUS-WIDE}(a), \textbf{MIRFlickr}(b) and \textbf{Pascal-Sentence}(c).}\label{conv}
\end{figure*}

\subsection{Results}
Based on the preceding settings, we test the performance of the proposed DS²L on the cross-modal retrieval. Experiments are performed for two typical retrieval tasks, i.e.,image querying text and text querying image, which are respectively abbreviated as 'I2T' and 'T2I' in this paper. The meaning of I2T is as follows: Given an image as a query, the model tries to retrieve relevant text modality information. Similarly, T2I represents the task of retrieving the relevant images from the image database when given a text. In this section, we report the results of MAP comparison on NUS-WIDE and MIRFlickr. The best performance is marked by bold. As shown in Table \uppercase\expandafter{\romannumeral3}, we can observe that our proposed DS²L obtains the best performance among all the compared methods. Specifically, DS²L achieves average improvements of $3.21\%$ and $1.73\%$ over the best baselines on NUS-WIDE and MIRFlickr respectively.

We further discuss the comparative MAP results as follows. As introduced before, CCA studies the correlation coefficients to maximize the feature correlation among different modalities. KCCA and ml-CCA are two extensions of CCA. KCCA uses a nonlinear mapping to map the original feature into a shared subspace, while ml-CCA learns a shared subspace with multi-label annotations. Unlike these CCA-based approaches which learn a shared subspace via maximizing the feature correlations among multi-modal data, KDM maximizes the kernel dependence to learn subspace representation.  While among CCA, KCCA, ml-CCA and KDM, KDM achieves the best performance relatively. We can infer that the main reason leading to this is that KDM exploits the consistency between feature-similarity and semantic-similarity for cross-modal retrieval. Compared to KDM, CKD constructs a semantic graph to the learning model, which preserves both the semantic structure and the dependence among multiple modalites. This contributes to the excellence of CKD. Although CKD has shown its superior performance for cross-modal retrieval, its performance is limited because of not making full use of discriminative semantically structural information. To overcome this problem, our DS²L exploit the label information to build up a similarity preservation term. This term is expected to compensate for the shortcomings of insufficient use of discriminative information and make the learned representation more discriminative. 

Our proposed DS²L mainly consists of three parts: kernel dependence maximization, discriminative structure preservation and similarity preservation. We conduct some ablation experiments on three benchmark datasets to verify the impact of each part on the retrieval performance. According to Eq.7, when we set $\alpha_1=\alpha2=0$, the experiment of DS²L($\alpha_1=\alpha2=0$) is conducted to show the effect of similarity preservation on KDM. DS²L($\beta=0$) only consider the semantically structural preservation for each modal. DS²L($\theta=0$) represents the performance of KDM. As illustrated in Table \uppercase\expandafter{\romannumeral4}, we can find that DS²L performs the best among DS²L($\alpha_1=\alpha2=0$), DS²L($\beta=0$) and DS²L($\theta=0$), which indicates the fact that integrating kernel dependence maximization, discriminative structure preservation and similarity preservation into a joint framework can lead to the most improvement on cross-modal retrieval performance. 


\subsection{Parameter sensitivity analysis}
We carry out experiments to show the influence of the parameters appeared in this paper on the retrieval performance. As Eq.7 shows, with $\beta$ fixed as 1, we have three parameters $\alpha_1$, $\alpha_2$ and $\theta$ to control the weight of the three terms. When exploring the impact of $\alpha_1$, $\alpha_2$, we fix the value of $\theta$ as $10^3$ and $10^2$ on NUS-WIDE and MIRFlickr,  respectively. Then we vary the value of $\alpha_1$, $\alpha_2$ in the candidate range of $\{10^{-5},10^{-4},10^{-3},10^{-2},10^{-1},1,10,10^{2},10^{3},10^{4},10^{5}\}$. As shown in Fig.~\ref{paracom1}, we can see that the DS²L can achieve stable MAP results in a large range on the three benchmark datasets which shows our model has weak dependence on $\alpha_1$ and $\alpha_2$. We can also infer that DS²L on Pascal-Sentence is more sensitive to $\alpha_1$ and $\alpha_2$ than on NUS-WIDE and MIRFlickr. In addition, we explore the impact of $\theta$ on cross-modal retrieval. We determine $\alpha_1= 1$ and $\alpha_2=10$, as well as $\alpha_1= 10^{-5}$ and $\alpha_2=1$ on NUS-WIDE and MIRFlickr, respectively. The result is shown in Fig.5. Furthermore, we tune the dimension d of Hilbert space from the range of $\{10,20,30,40,50,60\}$. Fig.~\ref{parad} displays the MAP scores of cross-modal retrieval versus different values of d. As can be seen in Fig.~\ref{parad}, when we set d as 60, 50 and 50 on NUS-WIDE, MIRFlickr and Pascal-Sentence, DS²L can obtain its best performance.

\subsection{Convergence experiment}
In the preceding section, we have theoretically analyzed the convergence behaviour of our algorithm. We reach the conclusion that our algorithm is decreasing monotonically, and will converge to the global minimum value. In this subsection, we conduct experiments to validate the convergence property. Specifically, we compute the value of the objective function in Eq.7. Fig.~\ref{conv} shows the relationship between the value of the objective function and the number of iteration on three datasets. In Fig.~\ref{conv}, the value of the objective function quickly tends to be stable as the number of iterations increases. This verifies the theoretical analysis before and shows the efficiency of the designed iterative optimization strategy in Algorithm 1 as well.

\section{Conclusion}\label{conclusion}
In this paper, we propose a discriminative supervised subspace learning method for cross-modal retrieval that combines kernel dependence maximization and discriminative semantically structural preservation into a unified framework. Unlike a set of the existing subspace learning approaches, a  similarity preservation term is constructed to compensate for the shortcomings of insufficient use of discriminative data and make the learned representation more discriminative. The DS²L is evaluated on three publicly available datasets, and the role of each component of our model is explored experimentally. Experimental results demonstrate the effectiveness of the proposed DS²L and the superiority compared with several classic subspace learning based methods.

\ifCLASSOPTIONcaptionsoff
  \newpage
\fi



\bibliographystyle{IEEEtran}
\bibliography{ref.bib}

\end{document}